\documentclass{article}

\PassOptionsToPackage{dvipsnames,table}{xcolor}
\usepackage[preprint]{neurips_2026}

\usepackage[utf8]{inputenc}
\usepackage[T1]{fontenc}
\usepackage{microtype}
\usepackage{graphicx}
\usepackage{pdflscape}
\usepackage{subcaption}
\usepackage{booktabs}
\usepackage{multirow}
\usepackage{makecell}
\usepackage{array}
\usepackage{adjustbox}
\usepackage{amsmath}
\usepackage{amssymb}
\usepackage{amsthm}
\usepackage{mathtools}
\usepackage{bm}
\usepackage{enumitem}
\usepackage{xcolor}
\usepackage{tikz}
\usetikzlibrary{arrows.meta,positioning,calc}
\usepackage{url}
\usepackage{hyperref}
\usepackage[capitalize,noabbrev]{cleveref}

\definecolor{retacoblue}{HTML}{3973AC}
\definecolor{retacoorange}{HTML}{E68A2E}
\definecolor{retacogray}{HTML}{D9D9D9}
\definecolor{retacogreen}{HTML}{4C956C}
\colorlet{retacorow}{retacoblue!15}
\hypersetup{colorlinks=true,linkcolor=retacoblue,citecolor=retacoblue,urlcolor=retacoblue}

\newcommand{\Vocab}{\mathcal{V}}
\newcommand{\Selected}{\mathcal{S}}
\newcommand{\Tail}{\mathcal{T}}
\newcommand{\E}{\mathbb{E}}

\newcommand{\Ind}{\mathbb{I}}
\newcommand{\KL}{D_{\mathrm{KL}}}
\newcommand{\BerKL}{d_{\mathrm{Ber}}}
\newcommand{\Coarse}{C_{\Selected}}
\newcommand{\qsel}{q^{\Selected}}
\newcommand{\psel}{p^{\Selected}}
\newcommand{\qtail}{q^{\Tail}}
\newcommand{\ptail}{p^{\Tail}}
\newcommand{\rbeta}{r_{\beta}}
\newcommand{\targ}{t_{\beta}}
\newcommand{\method}{ReTaCo}
\newcommand{\methodfull}{Residual-Target Control for On-Policy Distillation}
\newcommand{\RKLhat}{\widehat{D}_{\mathrm{RKL}}^{\mathrm{k3+}}}
\newcommand{\sg}{\operatorname{sg}}

\DeclareMathOperator{\softmax}{softmax}
\DeclareMathOperator{\logsumexp}{logsumexp}
\DeclareMathOperator{\TV}{TV}

\theoremstyle{plain}
\newtheorem{theorem}{Theorem}[section]
\newtheorem{proposition}[theorem]{Proposition}
\newtheorem{corollary}[theorem]{Corollary}

\theoremstyle{definition}

\theoremstyle{remark}

\hypersetup{pdftitle={\method: \methodfull}}
\title{\method: \methodfull}

\author{%
	\textbf{Zixiang Ni\textsuperscript{1,$*$}, Zhuo Hu\textsuperscript{2,$*$}, Renjie Cao\textsuperscript{3,4,$*$}, Weijie Ren\textsuperscript{2}, Binqin Shi\textsuperscript{1}, Weijia Zhang\textsuperscript{5},} \\
	\textbf{Shuheng Cao\textsuperscript{6}, Zhicheng Shi\textsuperscript{2}, Zhenhao Zhang\textsuperscript{7}, Haomin Wen\textsuperscript{8,$\dagger$}, Zhiyuan Hu\textsuperscript{9,$\dagger$}} \\
	\textsuperscript{1}Xi'an Jiaotong University \quad
	\textsuperscript{2}Zhejiang University \quad
	\textsuperscript{3}Georgia Institute of Technology \\
	\textsuperscript{4}Boston College \quad
	\textsuperscript{5}Yale University \quad
	\textsuperscript{6}University of California, San Diego \\
	\textsuperscript{7}Tsinghua University \quad
	\textsuperscript{8}Shanghai Innovation Institute \quad
	\textsuperscript{9}Massachusetts Institute of Technology
}

\makeatletter
\newcommand{\@authornotes}{%
	\begingroup
	\renewcommand{\thefootnote}{\fnsymbol{footnote}}%
	\footnotetext[1]{Equal contribution.}%
	\footnotetext[2]{Corresponding authors.}%
	\endgroup}
\newcommand{\authornotes}{%
	\@ifundefined{if@submission}{\@authornotes}{\if@submission\else\@authornotes\fi}}
\makeatother

\begin{document}
	\maketitle
	\authornotes
	
	\begin{abstract}
		
		On-policy distillation trains a student on its own generated prefixes with token-level teacher feedback, but transmitting or storing the teacher's full-vocabulary distribution at every generated token is costly. Entropy-aware on-policy distillation (EOPD) supplements the reverse Kullback--Leibler (KL) divergence with forward supervision that helps the student recover plausible tokens it underestimates; to limit cost, this supervision uses only the teacher's top-$k$ tokens. Because EOPD renormalizes the retained probabilities, its target assigns all mass to the selected tokens and none to the omitted vocabulary. We prove that the resulting loss keeps pushing the student's selected mass toward one even after the student matches the teacher's relative probabilities within the selected set. Consequently, the teacher distribution itself is not a stationary point whenever the omitted vocabulary has positive teacher probability. Motivated by this finding, we propose \method{} (\methodfull). Its forward target keeps the top-$k$ tokens individually and groups all remaining tokens into a single residual symbol; \method{} pairs this target with a single-sample estimator whose expectation equals the full-vocabulary reverse KL. Given the teacher's selected mass $m$ and a parameter $\beta\in[0,1]$, the residual target is $(1-\beta)(1-m)$, and the relative probabilities of the selected tokens are unchanged: $\beta=0$ preserves the teacher's mass, and larger $\beta$ moves more target mass onto the selected tokens. At a fixed prefix, we prove that the population objective has a unique optimum whose selected mass lies between $m$ and $m+\beta(1-m)$ and increases monotonically with $\beta$. At $\beta=0$, the forward term still provides non-vanishing recovery gradients for underestimated selected tokens. Numerical optimization confirms the predicted optimal mass and conditional distributions, and across three teacher--student pairs, \method{} outperforms EOPD on most mathematics and code benchmarks.
		
		
	\end{abstract}

	\section{Introduction}
	\label{sec:introduction}
	
	Transferring the capabilities of large language models to smaller models is an important step toward making these capabilities accessible under limited computational resources. On-policy distillation (OPD) addresses this goal by training a student on its own generated prefixes, with token-level supervision from a stronger teacher~\citep{agarwal2024policy,lu2025onpolicydistillation}. Building on earlier approaches to knowledge distillation~\citep{hinton2015distilling,kim2016sequence}, OPD obtains teacher feedback on the contexts that the student actually encounters and thereby reduces the mismatch between training and inference. However, transmitting the teacher's full-vocabulary probabilities at every generated token incurs substantial communication costs. Sending only the teacher's top-$k$ tokens and their probabilities reduces this cost, but leaves open how the training objective should treat the omitted probability mass.
	
	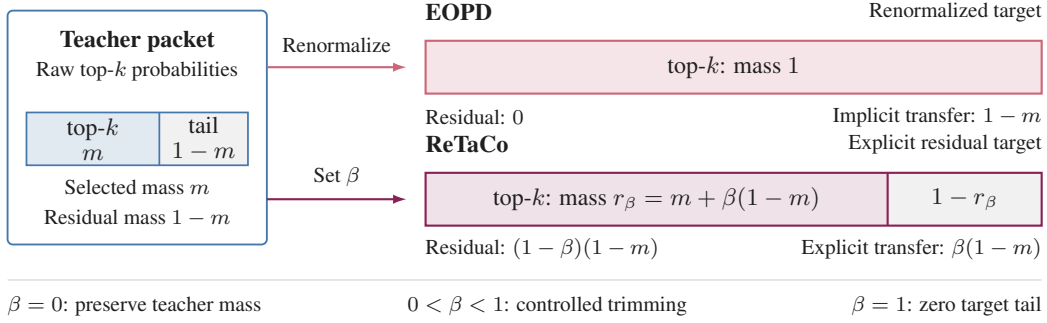
\begin{figure}[t]
		\centering
\definecolor{overviewblue}{HTML}{4477AA}
\definecolor{overviewrose}{HTML}{CC6677}
\definecolor{overviewwine}{HTML}{882255}
\definecolor{overviewneutral}{HTML}{B0B0B0}
\begin{tikzpicture}[
    x=1cm,y=1cm,
    font=\fontsize{9}{10.5}\selectfont,
    text=black!88,
    title/.style={font=\fontsize{9}{10.5}\selectfont\bfseries},
    note/.style={font=\fontsize{8}{9.5}\selectfont},
    arrow/.style={-{Latex[length=1.8mm,width=1.25mm]},line width=0.8pt},
    every node/.style={inner sep=0pt,outer sep=0pt}
]
    \draw[overviewblue,line width=0.7pt,rounded corners=2pt]
        (0.02,0.35) rectangle (3.45,3.45);
    \node[title] at (1.735,3.05) {Teacher packet};
    \node[note] at (1.735,2.65) {Raw top-$k$ probabilities};

    \fill[overviewblue!16] (0.27,1.40) rectangle (2.028,2.10);
    \fill[overviewneutral!18] (2.028,1.40) rectangle (3.20,2.10);
    \draw[overviewblue,line width=0.65pt] (0.27,1.40) rectangle (3.20,2.10);
    \draw[overviewblue,line width=0.65pt] (2.028,1.40) -- (2.028,2.10);
    \node[align=center] at (1.149,1.75) {top-$k$\\[-1pt]$m$};
    \node[align=center] at (2.614,1.75) {tail\\[-1pt]$1-m$};
    \node[note,align=center] at (1.735,0.90) {Selected mass $m$\\[2pt]Residual mass $1-m$};

    \draw[arrow,overviewrose] (3.45,2.70) -- (5.30,2.70);
    \node[note] at (4.375,3.00) {Renormalize};
    \node[title,anchor=west] at (5.55,3.43) {EOPD};
    \node[note,anchor=east] at (13.70,3.43) {Renormalized target};
    \fill[overviewrose!17] (5.55,2.35) rectangle (13.70,3.05);
    \draw[overviewrose,line width=0.8pt] (5.55,2.35) rectangle (13.70,3.05);
    \node at (9.625,2.70) {top-$k$: mass $1$};
    \node[note,anchor=west] at (5.55,2.04) {Residual: $0$};
    \node[note,anchor=east] at (13.70,2.04) {Implicit transfer: $1-m$};

    \draw[arrow,overviewwine] (3.45,0.95) -- (5.30,0.95);
    \node[note] at (4.375,1.25) {Set $\beta$};
    \node[title,anchor=west] at (5.55,1.68) {\method};
    \node[note,anchor=east] at (13.70,1.68) {Explicit residual target};
    \fill[overviewwine!13] (5.55,0.60) rectangle (11.6625,1.30);
    \fill[overviewneutral!18] (11.6625,0.60) rectangle (13.70,1.30);
    \draw[overviewwine,line width=0.8pt] (5.55,0.60) rectangle (13.70,1.30);
    \draw[overviewwine,line width=0.8pt] (11.6625,0.60) -- (11.6625,1.30);
    \node at (8.60625,0.95) {top-$k$: mass $\rbeta=m+\beta(1-m)$};
    \node at (12.68125,0.95) {$1-\rbeta$};
    \node[note,anchor=west] at (5.55,0.29) {Residual: $(1-\beta)(1-m)$};
    \node[note,anchor=east] at (13.70,0.29) {Explicit transfer: $\beta(1-m)$};

    \draw[overviewneutral!60,line width=0.4pt] (0.02,-0.12) -- (13.70,-0.12);
    \node[note,anchor=west] at (0.02,-0.47) {$\beta=0$: preserve teacher mass};
    \node[note] at (7.16,-0.47) {$0<\beta<1$: controlled trimming};
    \node[note,anchor=east] at (13.70,-0.47) {$\beta=1$: zero target tail};
\end{tikzpicture}
		\caption{\textbf{Residual probability as an explicit forward target.}
			The teacher's top-$k$ probabilities carry selected mass $m$, leaving residual mass $1-m$.
			EOPD renormalizes the selected probabilities to sum to one, assigning zero target mass to the residual.
			\method{} instead sets the residual target to $(1-\beta)(1-m)$ and transfers $\beta(1-m)$ to the selected tokens.
			Both targets preserve the teacher's relative probabilities within the selected set. Bar widths are schematic.}
		\label{fig:overview}
	\end{figure}
	
	OPD typically minimizes a per-token reverse KL divergence, which can be estimated from the sampled token alone. However, reverse KL provides only a weak signal for recovering teacher-supported tokens that the student rarely samples. Entropy-aware on-policy distillation (EOPD)~\citep{jin2026eopd} addresses this problem by adding forward KL at positions with high teacher entropy, using renormalized top-$k$ teacher probabilities. This renormalization preserves the relative probabilities of the selected tokens, but it inflates their total probability to one. When evaluated against the student's full-vocabulary probabilities, the resulting loss encourages the student to recover the selected tokens and also to move probability away from the remaining vocabulary.
	
	To make this effect precise, let $q$ and $p$ denote the teacher and student next-token distributions. For the teacher's top-$k$ set $\Selected$, let $m=q(\Selected)$ and $P=p(\Selected)$ denote the total probabilities that the teacher and student assign to this set. We show that the practical forward term decomposes exactly as
	\begin{equation}
		\mathcal{L}_{\mathrm{ren}}
		= \KL(\qsel\|\psel)-\log P,
		\label{eq:intro_identity}
	\end{equation}
	where $\qsel$ and $\psel$ are the teacher and student distributions conditioned on the selected set, which we call their conditional shapes. Even after these conditional distributions match, $-\log P$ continues to push the student's selected mass upward. Consequently, the teacher distribution itself is not a stationary point of this loss when $m<1$. Reverse KL moderates this pressure but does not remove it.
	
	This decomposition suggests making residual probability an explicit part of the supervision target. We introduce \method{}, short for \textbf{Re}sidual-\textbf{Ta}rget \textbf{Co}ntrol, which specifies how much probability the forward target keeps outside the selected set. We group the remaining tokens into one residual symbol and assign it probability $(1-\beta)(1-m)$. This gives the selected-mass target
	\begin{equation}
		\rbeta=m+\beta(1-m), \qquad \beta\in[0,1].
		\label{eq:intro_target}
	\end{equation}
	Here, $\beta=0$ preserves the teacher's mass allocation, whereas $\beta>0$ transfers a fraction $\beta$ of the tail mass to the selected set. The conditional shape within the selected set remains unchanged. Figure~\ref{fig:overview} illustrates this allocation.
	
	Our contributions are threefold.
	\begin{itemize}[leftmargin=1.35em,itemsep=0.25em,topsep=0.25em]
		\item We show that renormalized top-$k$ forward KL implicitly sets the selected-mass target to one. We prove that the teacher is not a stationary point of this loss and characterize the equilibrium when this loss is combined with reverse KL.
		\item We introduce \method{} with an explicit residual target and $O(k)$ teacher communication. At a fixed prefix, we characterize the unique population optimum, bound its selected mass, and prove that this mass increases monotonically with $\beta$. We also show that underestimated selected tokens still receive non-vanishing recovery gradients at $\beta=0$.
		\item We numerically verify the predicted mass and conditional distributions, evaluate downstream performance across three model pairs, and characterize the effects of residual targets and objective components through ablations on Qwen3-1.7B trained for one epoch.
	\end{itemize}


	\section{Preliminaries}
	\label{sec:preliminaries}

	\subsection{On-Policy Distillation and Selected Support}
	
	\paragraph{On-policy token distributions.}
	OPD queries a teacher at prefixes generated by the student~\citep{agarwal2024policy,gu2024minillm,lu2025onpolicydistillation}. At such a prefix $x$, let $q=(q_i)_{i\in\Vocab}$ and $p=(p_i)_{i\in\Vocab}$ be the teacher and student next-token distributions over the vocabulary $\Vocab$. We study the ideal per-prefix reverse-KL objective $\KL(p\|q)$ with $x$ and $q$ fixed. This isolates token-level supervision from rollout sampling, stale policies, and clipping in the practical training loop (\cref{sec:limitations}).
	
	\paragraph{Selected mass and conditional shape.}
	Let $\Selected$ contain the teacher's top-$k$ tokens, and let $\Tail$ denote its complement in $\Vocab$. The selected teacher and student masses are
	\begin{equation}
		m=q(\Selected)=\sum_{i\in\Selected}q_i,
		\qquad
		P=p(\Selected)=\sum_{i\in\Selected}p_i .
		\label{eq:masses}
	\end{equation}
	For $i\in\Selected$ and $j\notin\Selected$, write
	\begin{equation}
		q_i^{\Selected}=\frac{q_i}{m},\quad
		p_i^{\Selected}=\frac{p_i}{P},\qquad
		q_j^{\Tail}=\frac{q_j}{1-m},\quad
		p_j^{\Tail}=\frac{p_j}{1-P}.
	\end{equation}
	We consider teacher and student distributions with full support, as produced by finite-logit softmax models, and assume $0<m,P<1$ in the derivations. Endpoint identities are interpreted by continuity wherever the corresponding limits are well defined. Let $\BerKL(u\|v)=u\log(u/v)+(1-u)\log[(1-u)/(1-v)]$ denote the Bernoulli KL divergence. Conditioning on the selected and residual sets gives
	\begin{align}
		\KL(q\|p)
		&=
		\BerKL(m\|P)
		+m\KL(\qsel\|\psel)
		+(1-m)\KL(\qtail\|\ptail), \label{eq:fkl_chain}\\
		\KL(p\|q)
		&=
		\BerKL(P\|m)
		+P\KL(\psel\|\qsel)
		+(1-P)\KL(\ptail\|\qtail). \label{eq:rkl_chain}
	\end{align}
	The Bernoulli term measures disagreement in mass allocation, whereas the remaining terms measure disagreement within each set. This separation lets us distinguish recovering selected tokens from changing their total probability.
	
	\subsection{Entropy-Aware On-Policy Distillation}
	\label{sec:eopd}
	
	EOPD supplements reverse KL with forward KL to recover teacher-supported tokens that the student underestimates~\citep{jin2026eopd}. The motivation follows from the ideal reverse-KL logit gradient. For student logits $z$ with $p=\softmax(z)$,
	\[
		\frac{\partial\KL(p\|q)}{\partial z_j}
		=p_j\left(\log\frac{p_j}{q_j}-\KL(p\|q)\right).
	\]
	For a fixed full-support teacher, this gradient vanishes as $p_j\to0$, whereas the full forward-KL gradient $p_j-q_j$ approaches $-q_j$. Forward supervision can therefore supply a recovery signal even when the student rarely samples a teacher-supported token.
	
	EOPD applies its forward term at positions with high teacher entropy and uses only the teacher's top-$k$ probabilities to limit communication. Let $H(q)=-\sum_{i\in\Vocab}q_i\log q_i$ be the teacher entropy and $g_{\mathrm{ent}}(x)=\Ind[H(q)>\tau]$ the gate at prefix $x$, where $\tau$ is an entropy threshold and $\Ind$ is the indicator function. With a forward weight $\lambda>0$, the gated token-level contribution is $\lambda g_{\mathrm{ent}}(x)\mathcal{L}_{\mathrm{ren}}$. When the student's probabilities are normalized over the full vocabulary, EOPD's forward loss is
	\begin{equation}
		\mathcal{L}_{\mathrm{ren}}(q,p;\Selected)
		=
		\sum_{i\in\Selected}q_i^{\Selected}
		\log\frac{q_i^{\Selected}}{p_i}.
		\label{eq:ren_fkl}
	\end{equation}
	The teacher probabilities are normalized within $\Selected$, whereas the denominator $p_i$ is normalized over the full vocabulary. EOPD combines this gated forward contribution with a proximal policy optimization (PPO)-style reverse update. If the student is also normalized within $\Selected$, the loss instead becomes $\KL(\qsel\|\psel)$; the mass effect below specifically concerns \cref{eq:ren_fkl}.

	\section{Method}
	\label{sec:method}
	
	\method{} preserves EOPD's recovery signal for the selected tokens but makes the residual target explicit. We first show that EOPD's renormalized forward term implicitly targets unit selected mass (\cref{sec:analysis}). We then construct the residual target (\cref{sec:target}) and the combined objective (\cref{sec:sampled_objective}), characterize its population optimum and recovery gradients (\cref{sec:population}), and describe the implementation (\cref{sec:implementation}).

	\subsection{The Renormalization Effect in EOPD}
	\label{sec:analysis}
	
	EOPD's forward term supervises the selected tokens, but its asymmetric normalization also changes the target mass. Factoring the student probabilities into selected mass and conditional shape exposes this second effect.
	
	\begin{proposition}[Hidden selected-mass target]
		\label{prop:hidden_mass}
		For any teacher-selected set $\Selected$ with student mass $P>0$,
		\begin{equation}
			\mathcal{L}_{\mathrm{ren}}(q,p;\Selected)
			=
			\KL(\qsel\|\psel)-\log P.
			\label{eq:hidden_mass}
		\end{equation}
		Hence the conditional term is minimized at $\psel=\qsel$, whereas the mass term is minimized at $P=1$.
	\end{proposition}
	
	Substituting $p_i=Pp_i^{\Selected}$ proves the identity. Equation~\eqref{eq:ren_fkl} can also be read as forward KL from the full-vocabulary target $(\qsel,0_{\Tail})$ to the student, where $0_{\Tail}$ assigns zero probability to every tail token. Intuitively, the target places no mass on the tail, but the student is normalized over the full vocabulary, so any tail mass the student keeps is penalized through $-\log P$. Even when the conditional shapes match, $-\log P$ continues to favor larger selected mass. The same effect is visible in the logit gradient.
	
	\begin{corollary}[Teacher matching is not stationary]
		\label{cor:not_stationary}
		Let $p=\softmax(z)$. The logit gradient of \cref{eq:ren_fkl} is
		\begin{equation}
			\frac{\partial\mathcal{L}_{\mathrm{ren}}}{\partial z_j}
			=
			\begin{cases}
				p_j-q_j^{\Selected}, & j\in\Selected,\\
				p_j, & j\notin\Selected.
			\end{cases}
			\label{eq:ren_gradient}
		\end{equation}
		At $p=q$ and $m<1$, every residual logit has a positive gradient and decreases under direct gradient descent on the logits.
	\end{corollary}
	
	The entropy gate does not remove this normalization effect at active prefixes: it determines \emph{where} supervision applies, whereas $m$ determines the reassigned mass $1-m$. In general, entropy does not determine top-$k$ mass, so equal-entropy prefixes need not receive equal mass shifts.
	
	\paragraph{Equilibrium with reverse KL.}
	To isolate whether the reverse term removes this mass shift, replace the practical reverse update by ideal reverse KL and hold an active prefix fixed. In this per-prefix comparison, $\alpha=\lambda>0$ because $g_{\mathrm{ent}}(x)=1$. The idealized joint objective is
	\begin{equation}
		\mathcal{J}_{\mathrm{ren}}(p)
		=\KL(p\|q)+\alpha\mathcal{L}_{\mathrm{ren}}(q,p;\Selected),
		\qquad \alpha>0.
		\label{eq:joint_ren}
	\end{equation}
	For fixed $P$, all conditional KL terms are minimized at the teacher conditionals. The remaining objective is $\BerKL(P\|m)-\alpha\log P$, so its optimum directly quantifies the competition between mass matching and the unit-mass forward target.
	
	\begin{proposition}[Mass inflation under the renormalized target]
		\label{prop:ren_equilibrium}
		The selected-mass optimum of \cref{eq:joint_ren} is the unique $P^*\in(0,1)$ satisfying
		\begin{equation}
			\log\frac{P^*(1-m)}{m(1-P^*)}
			=\frac{\alpha}{P^*}.
			\label{eq:ren_equilibrium}
		\end{equation}
		For every $m\in(0,1)$, $P^*>m$. As $m\to1$, the residual student mass obeys
		$1-P^*=e^{-\alpha}(1-m)+o(1-m)$.
	\end{proposition}

	Even ideal reverse KL therefore leaves selected-mass inflation. We next construct a forward target that still provides a recovery signal for the selected tokens and explicitly controls the residual mass.

	\subsection{Residual Representation and Target Construction}\label{sec:target}
	
	\paragraph{Representing residual probability.}
	To specify residual mass without transmitting individual tail probabilities, we group all tokens outside $\Selected$ into one residual symbol. The residual symbol exists only in the loss; the student still predicts over the full vocabulary $\Vocab$. Denote the aggregation by $\Coarse$, which gives
	\begin{equation}
		\Coarse p=\big(\{p_i\}_{i\in\Selected},\,1-P\big),
		\qquad
		\Coarse q=\big(\{q_i\}_{i\in\Selected},\,1-m\big).
		\label{eq:coarse_map}
	\end{equation}
	This representation contains the individual selected-token probabilities and the aggregate tail mass. By the reverse-KL chain rule,
	\begin{align}
		\KL(p\|q)
		=
		\KL(\Coarse p\|\Coarse q)
		+(1-P)\KL(\ptail\|\qtail),
		\label{eq:coarse_observable}\\
		\KL(\Coarse p\|\Coarse q)
		=
		\BerKL(P\|m)+P\KL(\psel\|\qsel).
		\label{eq:coarse_reverse}
	\end{align}
	Aggregation preserves the mass of the tail but omits its conditional divergence. We therefore use the coarse representation for the forward term and rely on sampled reverse KL (\cref{sec:sampled_objective}) for sensitivity to individual tail tokens.
	
	\paragraph{Controlling the residual target.}
	Choose $\beta\in[0,1]$ and keep a fraction $1-\beta$ of the teacher's residual mass, so that $1-\rbeta=(1-\beta)(1-m)$. Because the target sums to one, its selected mass is $\rbeta$; preserving the teacher's conditional shape then gives
	\begin{equation}
		\rbeta=m+\beta(1-m),\qquad
		\targ=
		\left(
		\left\{\rbeta\frac{q_i}{m}\right\}_{i\in\Selected},
		1-\rbeta
		\right),
		\quad \beta\in[0,1].
		\label{eq:mass_target}
	\end{equation}
	This target preserves the teacher's conditional shape and assigns an explicit probability to the residual. Its forward KL decomposes as
	\begin{equation}
		\KL(\targ\|\Coarse p)
		=
		\BerKL(\rbeta\|P)+\rbeta\KL(\qsel\|\psel).
		\label{eq:coarse_forward}
	\end{equation}
	The Bernoulli term supervises the selected mass, and the conditional term supervises the conditional shape of the selected tokens. At $\beta=0$, $\targ=\Coarse q$ preserves the teacher's mass; positive $\beta$ transfers $\beta(1-m)$ to the selected tokens, with zero residual target at $\beta=1$. Changing $\beta$ also changes the conditional coefficient $\rbeta$, coupling mass supervision with conditional weighting.

	\subsection{Reverse-KL Estimation and the Combined Objective}
	\label{sec:sampled_objective}
	
	The residual target supervises the aggregate tail mass but leaves the conditional shape of the tail unspecified. We complement it with a single-sample estimator whose expectation equals the full-vocabulary reverse KL. The estimator uses the teacher probability of a student-sampled token that may lie outside $\Selected$.
	
	\paragraph{Sampled reverse KL.}
	Let $y\sim p$ be the rollout token and $a_y=\log p_y-\log q_y$. With $\sg(\cdot)$ denoting stop-gradient, we use the following straight-through estimator, whose forward value is the standard low-variance estimate:
	\begin{equation}
		\RKLhat(p\|q;y)
		=
		\sg\!\left(e^{-a_y}+a_y-1\right)
		+\frac{a_y^2}{2}-\sg\!\left(\frac{a_y^2}{2}\right).
		\label{eq:k3plus}
	\end{equation}
	The estimator evaluates to $e^{-a_y}+a_y-1$ in the forward pass and uses the derivative of $a_y^2/2$ in the backward pass. Let $\theta$ denote the student parameters. At a fixed prefix with a fixed teacher, sampling from the current student without numerical clipping gives
	\begin{equation}
		\E_{y\sim p}\!\left[\RKLhat(p\|q;y)\right]=\KL(p\|q),
		\qquad
		\E_{y\sim p}\!\left[\nabla_\theta\RKLhat(p\|q;y)\right]
		=\nabla_\theta\KL(p\|q).
		\label{eq:k3plus_expectation}
	\end{equation}
	The sampled token is detached during differentiation, and the estimator requires only its teacher log-probability. Neither expectation differentiates through prefix sampling.
	
	\paragraph{Objective.}
	At each on-policy token, \method\ uses
	\begin{equation}
		\mathcal{L}_{\mathrm{sampled\text{-}ReTaCo}}(y)
		=
		\RKLhat(p\|q;y)
		+\alpha\KL(\targ\|\Coarse p)
		\label{eq:retaco}
	\end{equation}
	with a forward weight $\alpha\ge0$. The default training loss averages this objective over all unmasked response tokens without an entropy gate. Taking the expectation over a token sampled from the current student gives the ideal population objective at a fixed prefix, without clipping:
	\begin{equation}
		\mathcal{J}_{\mathrm{ReTaCo}}(p)
		=
		\KL(p\|q)+\alpha\KL(\targ\|\Coarse p).
		\label{eq:mass_population}
	\end{equation}
	At $\beta=0$, the forward target preserves the teacher's mass and conditional shape, and the reverse estimator equals the full-vocabulary reverse KL in expectation. Setting $\beta>0$ deliberately trims the tail. At $\beta=1$, the forward term equals \cref{eq:ren_fkl}; this setting reproduces EOPD's forward target at active prefixes, but not its entropy gate or PPO updates. When $k=|\Vocab|$, the residual vanishes and \cref{eq:mass_population} reduces to full-vocabulary reverse KL plus $\alpha$ times forward KL.

	\subsection{Population Optimum and Token Recovery}\label{sec:population}
	
	We now determine how the forward target changes the optimum of the combined objective. For fixed selected mass $P$, the shape-dependent terms in \cref{eq:mass_population} are
	\begin{equation}
		P\KL(\psel\|\qsel)
		+(1-P)\KL(\ptail\|\qtail)
		+\alpha\rbeta\KL(\qsel\|\psel).
		\label{eq:conditional_objective}
	\end{equation}
	All three terms vanish simultaneously only at $\psel=\qsel$ and $\ptail=\qtail$. The remaining scalar objective is $g(P)=\BerKL(P\|m)+\alpha\BerKL(\rbeta\|P)$. Because $g$ balances two terms, the target mass $\rbeta$ and the optimum mass $P^*$ need not coincide.
	
	\begin{theorem}[Unique and controllable selected-mass optimum]
		\label{thm:mass_equilibrium}
		Let $m\in(0,1)$, $\alpha>0$, and $\beta\in[0,1]$. The population objective in \cref{eq:mass_population} has a unique full-support minimizer. Its conditional shapes are $\psel=\qsel$ and $\ptail=\qtail$, and its selected mass $P^*$ is the unique root of
		\begin{equation}
			F(P)=
			\log\frac{P(1-m)}{m(1-P)}
			+\alpha\frac{P-\rbeta}{P(1-P)}
			=0.
			\label{eq:mass_stationarity}
		\end{equation}
		Moreover,
		\begin{equation}
			m\le P^*\le\rbeta,
			\qquad
			\frac{\mathrm{d}P^*}{\mathrm{d}\beta}>0
			\quad\text{for }\beta\in(0,1),
			\label{eq:mass_bounds}
		\end{equation}
		with positive one-sided derivatives at the endpoints. For $\beta=0$, $P^*=m$ exactly, whereas for $\beta>0$, both inequalities are strict.
	\end{theorem}
	
	Reverse KL favors teacher mass $m$, while forward KL favors $\rbeta$; their balance places the optimum between these two values. The stationarity function is strictly increasing because
	\begin{equation}
		F'(P)=\frac{1}{P(1-P)}
		+\alpha
		\frac{(P-\rbeta)^2+\rbeta(1-\rbeta)}
		{P^2(1-P)^2}>0.
		\label{eq:mass_derivative}
	\end{equation}
	Together with the signs of $F$ at the endpoints, this establishes a unique root; \cref{app:mass_proofs} gives the complete proofs.
	
	\paragraph{Token recovery at $\beta=0$.}
	
	\Cref{prop:mass_gradient} gives the logit gradient of the forward term for general $\beta$ and specializes it to $\beta=0$.
	
	\begin{proposition}[Forward logit gradient]
		\label{prop:mass_gradient}
		For student logits $z$ with $p=\softmax(z)$,
		\begin{equation}
			\frac{\partial\KL(\targ\|\Coarse p)}{\partial z_j}
			=
			\begin{cases}
				p_j-\rbeta q_j^{\Selected}, & j\in\Selected,\\[2mm]
				p_j\dfrac{\rbeta-P}{1-P}, & j\notin\Selected.
			\end{cases}
			\label{eq:mass_gradient}
		\end{equation}
		At $\beta=0$ and $p_j\to0$ for a selected token, the gradient approaches $-q_j$, and the target selected mass remains $m$.
	\end{proposition}
	
	Hence, recovering the selected tokens does not require a unit selected-mass target. For $\beta>0$, we measure the deliberate departure from the aggregated teacher by the total variation distance $\TV$:
	\begin{equation}
		\TV(\targ,\Coarse q)=\beta(1-m).
		\label{eq:tv_shift}
	\end{equation}
	The transferred mass $\delta_\beta=\beta(1-m)$ measures target displacement. A fixed $\beta$ has a small absolute effect when the teacher's top-$k$ mass is near one and a larger effect when it is smaller.
	
	\subsection{Implementation}\label{sec:implementation}
	
	At each unmasked response position, the student rollout supplies $y$, and the teacher transmits $\log q_y$ with the top-$k$ token IDs and their \emph{raw} log-probabilities. Because they are normalized over the full vocabulary, these probabilities sum to $m$ rather than one. Together with the student's log-probabilities at the same IDs, they define $\targ$ and $\Coarse p$ for the forward loss, and $\log p_y$ and $\log q_y$ give the sampled reverse term in \cref{eq:k3plus}.
	
	The teacher transmits $O(k)$ values per token. Given full-vocabulary student log-probabilities, the additional loss computation is also $O(k)$. This reduces teacher--student communication, although the teacher still computes full-vocabulary softmax normalization and top-$k$ selection. \Cref{app:implementation} gives the per-token computation and numerical details.

\section{Experiments}
\label{sec:experiments}

We compare \method{} with two baselines on three teacher--student pairs and then use ablations on Qwen3-1.7B to examine the residual target and the components of the objective (\cref{sec:ablations}).

\subsection{Experimental Setup}

\paragraph{Benchmarks and metrics.}
We evaluate mathematical reasoning on MATH500~\citep{hendrycks2021measuring,lightman2023lets}, the English text-only subset of OlympiadBench~\citep{he2024olympiadbench}, AMC~\citep{li2024numinamath}, AIME24, and AIME25. HumanEval+ and MBPP+ assess code generation through the extended EvalPlus tests~\citep{liu2023evalplus,evalplus2023mbpp}. ARC-C, MMLU-Pro, and GPQA-Diamond assess out-of-domain (OOD) performance after mathematics distillation~\citep{clark2018arc,wang2024mmlupro,rein2023gpqa}; we do not claim that these benchmarks are absent from the models' pretraining data. We report avg@8 for mathematics, pass@1 for code, and accuracy on the OOD benchmarks using the protocols in \cref{app:benchmark_protocol}.

\paragraph{Models and comparisons.}
Table~\ref{tab:large_main} lists three teacher--student pairs. Sampled OPD and \method\ use the same direct $k3+$ reverse estimator. EOPD uses a renormalized forward term with an entropy gate and PPO-style reverse updates, so the comparison with EOPD measures the complete training procedures rather than the residual target alone.

\paragraph{Training and evaluation.}
Mathematics and code distillation use DAPO-Math-17k and the code subset of Eurus-2-RL-Data, respectively. Default \method\ uses $k=16$, $\alpha=1$, and $\beta=0$, and averages the loss over all response tokens without an entropy gate. Mathematics and OOD evaluations use the same epoch-1 checkpoints from mathematics distillation; code evaluations use separately code-distilled checkpoints. For mathematics, avg@8 is the mean correctness over eight completions per problem. \Cref{app:large_protocol} lists the optimization and decoding settings.

\begin{table}[t]
\centering
\caption{\textbf{Per-benchmark performance (\%).} We report avg@8 for mathematics, pass@1 for code, and accuracy for OOD benchmarks. OOD scores use the mathematics-distilled checkpoints. Within each teacher--student pair, \textbf{bold} marks the best score and \underline{underline} the second best; shaded rows are \method{}.}
\label{tab:large_main}
\small
\setlength{\tabcolsep}{3.2pt}
\setlength{\aboverulesep}{0pt}
\setlength{\belowrulesep}{0pt}
\renewcommand{\arraystretch}{1.2}
\begin{adjustbox}{max width=\linewidth}
\begin{tabular}{@{}l*{10}{c}@{}}
\toprule
& \multicolumn{5}{c}{Mathematics} & \multicolumn{2}{c}{Code} & \multicolumn{3}{c}{OOD} \\
\cmidrule(lr){2-6}\cmidrule(lr){7-8}\cmidrule(l){9-11}
Method & \makecell{MATH\\500} & \makecell{Olympiad\\Bench} & AMC & \makecell{AIME\\24} & \makecell{AIME\\25} & \makecell{Human\\Eval+} & MBPP+ & ARC-C & \makecell{MMLU-\\Pro} & \makecell{GPQA\\Diamond} \\
\midrule
\multicolumn{11}{@{}l}{\textit{Student: Qwen3-1.7B \quad Teacher: Qwen3-30B-A3B-Instruct-2507}} \\
Sampled OPD & \underline{82.40} & \underline{48.67} & \underline{52.11} & \underline{25.42} & \textbf{22.08} & 64.02 & \underline{52.38} & \textbf{82.51} & \underline{45.79} & \textbf{18.94} \\
EOPD & 79.80 & 45.54 & 48.04 & 21.67 & 17.92 & \underline{66.46} & 51.59 & 79.84 & 42.91 & 12.44 \\
\rowcolor{retacorow}
\method{} (ours) & \textbf{83.78} & \textbf{49.65} & \textbf{52.56} & \textbf{28.75} & \underline{21.67} & \textbf{68.29} & \textbf{53.17} & \underline{82.37} & \textbf{46.67} & \underline{17.49} \\
\midrule
\multicolumn{11}{@{}l}{\textit{Student: Qwen3.5-2B \quad Teacher: Qwen3.5-27B}} \\
Sampled OPD & \underline{71.73} & 40.07 & 45.63 & \underline{21.25} & \underline{17.50} & 48.17 & \underline{47.88} & 88.84 & 52.61 & 26.89 \\
EOPD & 70.93 & \underline{41.52} & \underline{46.08} & 20.00 & \underline{17.50} & \underline{50.61} & 44.18 & \underline{89.00} & \underline{53.73} & \underline{29.17} \\
\rowcolor{retacorow}
\method{} (ours) & \textbf{79.85} & \textbf{46.93} & \textbf{52.71} & \textbf{22.92} & \textbf{21.25} & \textbf{54.88} & \textbf{48.68} & \textbf{89.40} & \textbf{54.20} & \textbf{30.43} \\
\midrule
\multicolumn{11}{@{}l}{\textit{Student: Gemma4-E2B \quad Teacher: Gemma4-26B-A4B}} \\
Sampled OPD & 48.18 & 26.83 & 31.93 & 10.00 & 14.17 & \underline{75.61} & 66.38 & 62.18 & 28.99 & 13.26 \\
EOPD & \underline{62.50} & \underline{33.96} & \underline{34.19} & \underline{14.17} & \underline{16.67} & 74.39 & \underline{67.20} & \textbf{86.89} & \textbf{43.45} & \underline{13.64} \\
\rowcolor{retacorow}
\method{} (ours) & \textbf{79.47} & \textbf{47.93} & \textbf{56.17} & \textbf{22.50} & \textbf{19.58} & \textbf{76.83} & \textbf{67.79} & \underline{84.84} & \underline{41.60} & \textbf{31.00} \\
\bottomrule
\end{tabular}
\end{adjustbox}
\end{table}

\paragraph{Downstream performance.}
\label{sec:primary_comparison}
\method\ achieves higher point estimates than both baselines on all five mathematics tasks for Qwen3.5-2B and Gemma4-E2B, and on four for Qwen3-1.7B (Table~\ref{tab:large_main}). For example, on Qwen3.5-2B, \method\ raises MATH500 accuracy from 71.73\% (Sampled OPD) to 79.85\%. Sampled OPD remains slightly higher on Qwen3-1.7B AIME25. Across all three pairs, \method\ also leads both baselines on HumanEval+ and MBPP+.

\paragraph{Out-of-domain performance.}
OOD results are less uniform. For Qwen3.5-2B, \method\ leads on all three tasks. For Qwen3-1.7B, \method\ leads on MMLU-Pro but trails Sampled OPD on ARC-C and GPQA-Diamond. For Gemma4-E2B, \method\ leads on GPQA-Diamond, whereas EOPD leads on ARC-C and MMLU-Pro. 

\subsection{Residual Targets and Component Contributions}
\label{sec:ablations}

\begin{figure}[t]
    \centering
    \includegraphics[width=0.85\linewidth]{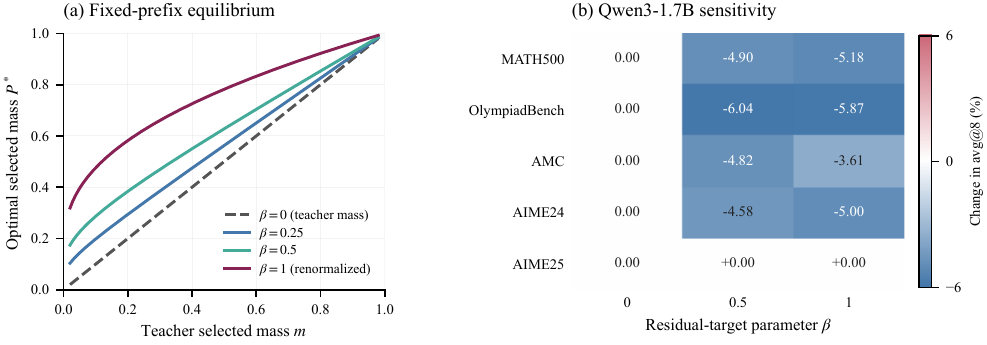}
    \caption{\textbf{Residual-target control in analysis and experiments.} (a) Selected-mass equilibria from \cref{eq:mass_stationarity} at $\alpha=1$; the dashed line denotes teacher mass. (b) Changes in Qwen3-1.7B avg@8 relative to $\beta=0$, computed from the residual-target rows in Table~\ref{tab:key_ablations}. Positive values indicate higher scores with trimming. All variants train for one epoch.}
    \label{fig:mass_equilibrium}
\end{figure}

\paragraph{Effect of the residual target.}
On Qwen3-1.7B, with all variants trained for one epoch, adding the default forward term to reverse-only training improves four of the five mathematics scores (Table~\ref{tab:key_ablations}), including MATH500 from 82.40\% to 83.78\%; only AIME25 decreases, from 22.08\% to 21.67\%. Preserving the teacher's mass ($\beta=0$) also outperforms partial or complete trimming on MATH500, OlympiadBench, and AMC. On AIME24, trimming reduces accuracy by 4.58 and 5.00 percentage points for $\beta=0.5$ and $\beta=1$, respectively, and AIME25 is unchanged (Figure~\ref{fig:mass_equilibrium}b). The benefit of residual trimming is therefore task-dependent. Because $\beta$ changes both the mass target and the conditional weight $\rbeta$, this comparison evaluates the target as a whole. The $\beta=1$ variant uses EOPD's forward target together with the same ungated direct reverse objective as the other ablations.

\begin{table}[t]
\centering
\caption{\textbf{Qwen3-1.7B target and component comparisons (avg@8, \%).} Let $R$ denote sampled reverse KL, $B_\beta=\BerKL(\rbeta\|P)$, and $C=\KL(\qsel\|\psel)$. The shaded row is the default, $R+B_0+mC$, and \textbf{bold} marks the best score in each column. All variants train for one epoch from the same initial student.}
\label{tab:key_ablations}
\small
\setlength{\tabcolsep}{6pt}
\setlength{\aboverulesep}{0pt}
\setlength{\belowrulesep}{0pt}
\renewcommand{\arraystretch}{1.2}
\begin{tabular}{@{}llccccc@{}}
\toprule
Setting & Objective & \makecell{MATH\\500} & \makecell{Olympiad\\Bench} & AMC & \makecell{AIME\\24} & \makecell{AIME\\25} \\
\midrule
\multicolumn{7}{@{}l}{\textit{Forward supervision and residual-target choice}} \\
\rowcolor{retacorow}
Full \method & $R+B_0+mC$ & \textbf{83.78} & \textbf{49.65} & 52.56 & \textbf{28.75} & 21.67 \\
No forward term & $R$ & 82.40 & 48.67 & 52.11 & 25.42 & \textbf{22.08} \\
Partial trimming ($\beta=0.5$) & $R+B_{0.5}+r_{0.5}C$ & 78.88 & 43.61 & 47.74 & 24.17 & 21.67 \\
Complete trimming ($\beta=1$) & $R+B_1+C$ & 78.60 & 43.78 & 48.95 & 23.75 & 21.67 \\
\midrule
\multicolumn{7}{@{}l}{\textit{Component variants under one-epoch training}} \\
Mass-only forward term & $R+B_0$ & 81.65 & 47.15 & 50.75 & 21.67 & \textbf{22.08} \\
Conditional-only forward term & $R+mC$ & 83.55 & 49.13 & \textbf{54.37} & 27.92 & 21.25 \\
No reverse term & $B_0+mC$ & 82.10 & 47.61 & 51.05 & 25.42 & 21.67 \\
\bottomrule
\end{tabular}
\end{table}

\paragraph{Mass and conditional supervision.}
Among the three component variants in Table~\ref{tab:key_ablations}, conditional-only forward supervision achieves the highest scores on MATH500, OlympiadBench, AMC, and AIME24, whereas mass-only supervision leads on AIME25. Relative to the full objective, removing the mass term lowers MATH500 from 83.78\% to 83.55\% and AIME24 from 28.75\% to 27.92\%, but raises AMC from 52.56\% to 54.37\%. Thus, the mass constraint does not uniformly improve downstream accuracy, and conditional recovery of the selected tokens explains most of the accuracy gains on these tasks.

\begin{figure}[t]
    \centering
    \includegraphics[width=0.85\linewidth]{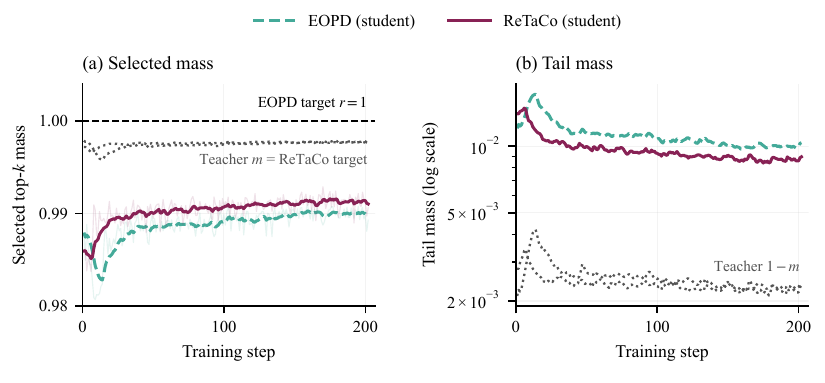}
    \caption{\textbf{Selected and residual mass during training.} Qwen3-1.7B student, Qwen3-30B-A3B-Instruct-2507 teacher, and $k=16$; EOPD and \method\ share the displayed interval of steps 1--202. (a) Student mass $P$ on the teacher's top-$k$ set for EOPD (teal, dashed) and \method\ (wine, solid). Gray dotted curves show teacher mass $m$ on each method's own rollouts; $m$ is also the \method\ forward target at $\beta=0$. The black dashed line marks EOPD's forward target $r=1$. (b) The same data expressed as tail mass $1-P$ and $1-m$, with a logarithmic vertical axis. Curves use exponential smoothing with span 11; faint traces in (a) show raw values.}
    \label{fig:training_dynamics}
\end{figure}

\paragraph{Selected and residual probability mass.}
\method\ brings the student's selected mass closer to the teacher's than EOPD does (Figure~\ref{fig:training_dynamics}). The comparison uses the Qwen3-1.7B student, the Qwen3-30B-A3B-Instruct-2507 teacher, and $k=16$, over the two runs' common interval of steps 1--202. Over the last 20 steps, the student's mean selected mass is $P=0.991252$ for \method\ and $P=0.989900$ for EOPD, and the teacher's is $m\approx0.9977$. The tail view shows the same improvement: \method\ assigns less excess probability outside the teacher's top-$k$ set than EOPD. Appendix~\ref{app:training_diagnostics} complements this comparison with policy-entropy trajectories for Sampled OPD and \method.

\paragraph{Numerical validation of the population objective.}
\label{sec:mechanism_validation}
To test the fixed-prefix predictions directly, we next examine the population objective. The equilibria in Figure~\ref{fig:mass_equilibrium}(a) lie between $m$ and $\rbeta$, with $P^*=m$ at $\beta=0$ (\cref{thm:mass_equilibrium}). For a uniform 80-token teacher with $k=16$ and $\alpha=1$, the renormalized target instead gives $P^*=0.5821$, far above $m=0.2$. Separately, joint optimization of a categorical student reaches the predicted mass and teacher conditionals, with a maximum $\ell_1$ error of $4.13\times10^{-8}$ in the conditional distribution over $\Selected$ (\cref{app:legacy_mechanism}).

	\section{Discussion and Limitations}
	\label{sec:limitations}
	
	At $\beta=0$, the forward objective preserves the teacher's aggregate mass and supervises each selected token individually, but it cannot resolve the conditional shape of the tail. Increasing $k$ refines this supervision at a higher communication cost. Sampled reverse KL remains sensitive to tail tokens in expectation. The equilibrium theorem holds at a fixed prefix under the population objective; it does not guarantee convergence with evolving prefixes, stale policies, or clipped updates.
	
The mathematics ablations show task-dependent effects of residual trimming and stronger forward supervision. The response-length experiment in \cref{app:legacy_stability} uses an entropy-gated forward term and a PPO-style reverse objective, so its results describe that variant rather than the default \method{} objective.
	

	\section{Conclusion}
	\label{sec:conclusion}
	
\method{} treats residual probability as an explicit supervision target. Its forward target over the top-$k$ tokens and a residual symbol preserves the teacher's conditional shape, and $\beta$ controls how much residual mass the target keeps. The single-sample reverse estimator equals the full-vocabulary reverse KL in expectation. The fixed-prefix population objective has a unique optimum whose selected mass increases monotonically with $\beta$, and at $\beta=0$ the objective still provides recovery gradients for the selected tokens. Numerical optimization matches these predictions. Experiments across three teacher--student pairs show higher point estimates on most mathematics and code benchmarks, with improvements across all three OOD benchmarks for Qwen3.5-2B. One-epoch component ablations indicate that conditional recovery explains most of the accuracy gains, whereas the mass term does not improve accuracy uniformly.
	

\clearpage
\subsection*{AI use statement}
Generative AI tools assisted with language polishing and editing of the manuscript and with figure preparation. The authors are responsible for the technical content, experimental results, references, and the final manuscript.

	\bibliography{references}
	\bibliographystyle{plainnat}
	
	\clearpage
	\appendix
	
	\section{Complete Derivations}
	\label{app:proofs}
	
	\subsection{KL chain rules}
	\label{app:chain_rule}
	
	Under the assumptions in \cref{sec:preliminaries}, substituting $q_i=mq_i^{\Selected}$ and $p_i=Pp_i^{\Selected}$ on the selected set yields
	\begin{align}
		\sum_{i\in\Selected}q_i\log\frac{q_i}{p_i}
		&=
		\sum_{i\in\Selected}mq_i^{\Selected}
		\left(
		\log\frac{m}{P}
		+\log\frac{q_i^{\Selected}}{p_i^{\Selected}}
		\right) \nonumber\\
		&=m\log\frac{m}{P}+m\KL(\qsel\|\psel).
		\label{eq:app_selected_fkl}
	\end{align}
	Applying the same expansion to the residual set gives
	\begin{equation}
		(1-m)\log\frac{1-m}{1-P}
		+(1-m)\KL(\qtail\|\ptail).
	\end{equation}
	Adding both contributions establishes \cref{eq:fkl_chain}, and exchanging $p$ and $q$ gives \cref{eq:rkl_chain}. Residual aggregation removes the conditional tail term but preserves its mass, as expressed in \cref{eq:coarse_reverse,eq:coarse_observable}.
	
	\subsection{The sampled reverse-KL estimator}
	
	Hold the prefix and teacher distribution fixed, and sample the token from the current student. For $y\sim p$ and $a_y=\log p_y-\log q_y$, the forward value of \cref{eq:k3plus} satisfies
	\begin{align}
		\E_{y\sim p}\!\left[e^{-a_y}+a_y-1\right]
		&=\sum_{i\in\Vocab}p_i
		\left(\frac{q_i}{p_i}+\log\frac{p_i}{q_i}-1\right)\\
		&=\KL(p\|q).
	\end{align}
	With the sampled token detached, the expected straight-through gradient is
	\begin{align}
		\E_{y\sim p}\!\left[\nabla_\theta\frac{a_y^2}{2}\right]
		&=\sum_{i\in\Vocab}p_i\log\frac{p_i}{q_i}\nabla_\theta\log p_i\\
		&=\nabla_\theta\KL(p\|q),
	\end{align}
	where the last equality uses $\sum_i p_i\nabla_\theta\log p_i=\nabla_\theta\sum_i p_i=0$. This proves \cref{eq:k3plus_expectation} for the ideal unclipped per-prefix estimator, without differentiation through prefix sampling.
	
	\subsection{The implicit target of renormalized forward KL}
	
	\paragraph{Proof of \cref{prop:hidden_mass}.}
	Within $\Selected$, the factorization $p_i=Pp_i^{\Selected}$ gives
	\begin{align}
		\mathcal{L}_{\mathrm{ren}}
		&=
		\sum_{i\in\Selected}q_i^{\Selected}
		\log\frac{q_i^{\Selected}}{Pp_i^{\Selected}} \nonumber\\
		&=
		\sum_{i\in\Selected}q_i^{\Selected}
		\log\frac{q_i^{\Selected}}{p_i^{\Selected}}
		-\log P
		=
		\KL(\qsel\|\psel)-\log P.
	\end{align}
	The conditional KL vanishes at teacher matching, whereas $-\log P$ decreases strictly on $(0,1]$ and is minimized at unit selected mass.
	\hfill$\square$
	
	\paragraph{Proof of \cref{cor:not_stationary}.}
	Ignoring teacher-only constants,
	\begin{equation}
		\mathcal{L}_{\mathrm{ren}}
		=-\sum_{i\in\Selected}q_i^{\Selected}\log p_i+\mathrm{const}.
	\end{equation}
	For $p=\softmax(z)$,
	$\partial\log p_i/\partial z_j=\Ind[i=j]-p_j$, where $\Ind$ is the indicator function. Hence
	\begin{align}
		\frac{\partial\mathcal{L}_{\mathrm{ren}}}{\partial z_j}
		&=
		-\sum_{i\in\Selected}q_i^{\Selected}
		\big(\Ind[i=j]-p_j\big) \nonumber\\
		&=
		\begin{cases}
			p_j-q_j^{\Selected},&j\in\Selected,\\
			p_j,&j\notin\Selected.
		\end{cases}
	\end{align}
	At $p=q$, every $j\notin\Selected$ has logit gradient $q_j>0$ under the full-support assumption. Teacher matching therefore fails to be stationary whenever residual mass is positive.
	\hfill$\square$
	
	\subsection{Equilibrium under the implicit unit-mass target}
	
	\paragraph{Proof of \cref{prop:ren_equilibrium}.}
	By \cref{eq:rkl_chain,eq:hidden_mass}, the conditional terms in \cref{eq:joint_ren} are
	\begin{equation}
		P\KL(\psel\|\qsel)
		+\alpha\KL(\qsel\|\psel)
		+(1-P)\KL(\ptail\|\qtail).
	\end{equation}
	For any fixed $P\in(0,1)$, they have the unique minimizer
	$\psel=\qsel$ and $\ptail=\qtail$. The remaining scalar objective is
	\begin{equation}
		g_{\mathrm{ren}}(P)
		=
		\BerKL(P\|m)-\alpha\log P.
	\end{equation}
	Its first two derivatives are
	\begin{align}
		g_{\mathrm{ren}}'(P)
		&=
		\log\frac{P(1-m)}{m(1-P)}-\frac{\alpha}{P},\\
		g_{\mathrm{ren}}''(P)
		&=
		\frac{1}{P(1-P)}+\frac{\alpha}{P^2}>0.
	\end{align}
	Because the derivative is strictly increasing and tends to $-\infty$ as $P\downarrow0$ and to $+\infty$ as $P\uparrow1$, it has exactly one root. Since
	$g_{\mathrm{ren}}'(m)=-\alpha/m<0$, this root satisfies $P^*>m$.
	
	For the asymptotic statement, let $\delta=1-m$ and $\epsilon=1-P^*$. Equation~\eqref{eq:ren_equilibrium} becomes
	\begin{equation}
		\log\frac{(1-\epsilon)\delta}{(1-\delta)\epsilon}
		=\frac{\alpha}{1-\epsilon}.
	\end{equation}
	Since the student residual is smaller than the teacher residual, both $\delta$ and $\epsilon$ approach zero as $m\to1$. Expansion gives
	$\log(\delta/\epsilon)=\alpha+o(1)$, and therefore
	$\epsilon=e^{-\alpha}\delta+o(\delta)$.
	\hfill$\square$
	
	\section{Proofs for \method}
	\label{app:mass_proofs}
	
	\subsection{Unique population optimum and monotonic control}
	
	\paragraph{Proof of \cref{thm:mass_equilibrium}.}
	Expanding the ideal population objective in \cref{eq:mass_population} gives
	\begin{align}
		\mathcal{J}_{\mathrm{ReTaCo}
		}
		&=
		\BerKL(P\|m)
		+\alpha\BerKL(\rbeta\|P) \nonumber\\
		&\quad+
		P\KL(\psel\|\qsel)
		+(1-P)\KL(\ptail\|\qtail)
		+\alpha\rbeta\KL(\qsel\|\psel).
		\label{eq:app_mass_expansion}
	\end{align}
	For fixed interior mass $P$, the conditional KL terms are non-negative and vanish simultaneously only at $\psel=\qsel$ and $\ptail=\qtail$. Optimizing the conditional shapes therefore reduces the population problem to
	\begin{equation}
		g(P)=\BerKL(P\|m)+\alpha\BerKL(\rbeta\|P).
	\end{equation}
	Direct differentiation yields
	\begin{equation}
		g'(P)
		=
		\log\frac{P(1-m)}{m(1-P)}
		+\alpha\frac{P-\rbeta}{P(1-P)}
		=F(P),
	\end{equation}
	and
	\begin{align}
		F'(P)
		&=
		\frac{1}{P(1-P)}
		+\alpha
		\frac{P^2-2P\rbeta+\rbeta}{P^2(1-P)^2}\nonumber\\
		&=
		\frac{1}{P(1-P)}
		+\alpha
		\frac{(P-\rbeta)^2+\rbeta(1-\rbeta)}
		{P^2(1-P)^2}>0.
	\end{align}
	Hence, $g$ is strictly convex. The limits $F(P)\to-\infty$ as $P\downarrow0$ and $F(P)\to+\infty$ as $P\uparrow1$ establish a unique interior root, including when $\rbeta=1$. The unique teacher conditionals then determine the full-support minimizer.
	
	If $\beta=0$, then $\rbeta=m$ and $F(m)=0$, so $P^*=m$. If $\beta>0$, then $\rbeta>m$ and
	\begin{equation}
		F(m)=\alpha\frac{m-\rbeta}{m(1-m)}<0.
	\end{equation}
	For $\rbeta<1$,
	\begin{equation}
		F(\rbeta)=
		\log\frac{\rbeta(1-m)}{m(1-\rbeta)}>0;
	\end{equation}
	when $\rbeta=1$, the limit as $P\uparrow1$ provides the corresponding positive upper sign. Since $F$ is strictly increasing, the root lies in $(m,\rbeta)$.
	
	Finally,
	\begin{equation}
		\frac{\partial F}{\partial\beta}
		=
		-\alpha\frac{1-m}{P(1-P)}.
	\end{equation}
	The implicit function theorem and $F'(P^*)>0$ give
	\begin{equation}
		\frac{\mathrm{d}P^*}{\mathrm{d}\beta}
		=
		\frac{\alpha(1-m)}
		{P^*(1-P^*)F'(P^*)}>0.
	\end{equation}
	At both parameter endpoints, the root remains interior and the denominator finite and positive, so the one-sided derivatives are also positive.
	\hfill$\square$
	
	\subsection{Forward gradient and intervention size}
	
	\paragraph{Proof of \cref{prop:mass_gradient}.}
	Up to a teacher-only constant,
	\begin{equation}
		\KL(\targ\|\Coarse p)
		=
		-\sum_{i\in\Selected}\rbeta q_i^{\Selected}\log p_i
		-(1-\rbeta)\log(1-P)+\mathrm{const}.
	\end{equation}
	For $j\in\Selected$, we have $\partial P/\partial z_j=p_j(1-P)$, whereas for $j\notin\Selected$, $\partial P/\partial z_j=-Pp_j$. Differentiating the two cross-entropy terms then gives
	\begin{equation}
		\frac{\partial\KL(\targ\|\Coarse p)}{\partial z_j}
		=
		\begin{cases}
			p_j-\rbeta q_j^{\Selected}, & j\in\Selected,\\[2mm]
			p_j(\rbeta-P)/(1-P), & j\notin\Selected.
		\end{cases}
	\end{equation}
	At $\beta=0$, we have $\rbeta=m$, so the first branch approaches $-mq_j^{\Selected}=-q_j$ as $p_j\to0$ for $j\in\Selected$. In this case, both branches are zero when $p=q$.
	\hfill$\square$
	
	\paragraph{Target displacement.}
	Let $\delta=\rbeta-m=\beta(1-m)$. Each selected target probability changes by
	$\delta q_i^{\Selected}$ relative to $\Coarse q$, and the residual probability changes by $-\delta$. Therefore
	\begin{equation}
		\TV(\targ,\Coarse q)
		=
		\frac{1}{2}
		\left(
		\sum_{i\in\Selected}\delta q_i^{\Selected}
		+\delta
		\right)
		=\delta.
	\end{equation}
	The $\ell_1$ change in the forward logit gradient relative to $\beta=0$ is similarly $2\delta$. The absolute changes in the selected branches sum to $\delta$, and those in the residual branches sum to
	$\sum_{j\notin\Selected}p_j\delta/(1-P)=\delta$. Thus, $\beta(1-m)$ directly characterizes the scale of the gradient intervention.
	
	\section{Reference Implementation}
	\label{app:implementation}
	
	\begin{table}[t]
		\centering
		\normalsize
		\setlength{\tabcolsep}{4.5pt}
		\caption{\textbf{Support and mass control across objectives.} The final column gives the mass preferred by the forward term alone. In the combined \method\ objective, reverse KL also influences the equilibrium.}
		\label{tab:objective_comparison}
		\begin{tabular}{lccc}
			\toprule
			Objective & Reverse information & Forward support & Forward mass \\
			\midrule
			Sampled reverse KL & Full KL in expectation & -- & -- \\
			Renormalized forward KL & -- & $k$ in full softmax & $P\!\to\!1$ \\
			Residual-target forward KL & -- & $k+1$ & $P\!\to\!\rbeta$ \\
			\method & Full KL in expectation & $k+1$ & $P\!\to\!\rbeta$ \\
			\bottomrule
		\end{tabular}
	\end{table}
	
	\subsection{Per-token computation}
	
	At each unmasked response position, we evaluate \cref{eq:retaco} as follows.
	\begin{enumerate}[leftmargin=1.6em,itemsep=0.25em]
		\item For the rollout token $y$, combine the student and teacher log-probabilities according to the straight-through estimator in \cref{eq:k3plus}.
		\item Apply log-softmax to the student logits and gather the $k$ entries indexed by the teacher-selected IDs. Recover $\log m=\logsumexp_{i\in\Selected}\log q_i$ and $\log P=\logsumexp_{i\in\Selected}\log p_i$, then compute both residual log-probabilities with a stable $\log(1-\exp(\cdot))$ routine.
		\item Form $\log\rbeta$ and the selected target log-probabilities
		$\log t_i=\log\rbeta+\log q_i-\log m$.
		\item Sum the selected and residual contributions for $\KL(\targ\|\Coarse p)$, add the sampled reverse term, and average over unmasked response tokens.
	\end{enumerate}
	The rollout token, teacher-selected IDs, raw top-$k$ log-probabilities, and set construction are detached. The student distribution remains a full-vocabulary softmax, so gradients from the aggregate tail probability also reach individual residual logits.
	
	\subsection{Numerical stability}

	If rounding makes selected mass indistinguishable from one, we rescale selected probabilities to sum to $1-\varepsilon$ and assign $\varepsilon$ to the residual. This preserves the conditional shape and normalization but perturbs the aggregate mass. We use $\varepsilon=10^{-7}$ in mixed-precision training and $\varepsilon=0$ in float64 checks.
	
	\section{Numerical Validation}
	\label{app:experiments}
	\label{app:legacy_mechanism}
	
	\subsection{Mass and Conditional-Shape Recovery}
	
	The synthetic validation jointly optimizes the selected mass and conditional distributions of a bimodal categorical student for $\beta\in\{0,0.5,1\}$. Table~\ref{tab:mass_recovery} compares the predicted and optimized student masses for teacher selected mass $m=0.6802847194$.

	\begin{table}[htbp]
        \centering
        \normalsize
        \caption{\textbf{Predicted and optimized selected mass.} The teacher selected mass is $m=0.6802847194$.}
        \label{tab:mass_recovery}
        \begin{tabular}{lrrr}
            \toprule
            Quantity & $\beta=0$ & $\beta=0.5$ & $\beta=1$ \\
            \midrule
            Predicted $P^*$ & 0.6802847194 & 0.7642681841 & 0.8703486945 \\
            Optimized $P$ & 0.6802847109 & 0.7642681719 & 0.8703486945 \\
            \bottomrule
        \end{tabular}
    \end{table}

	The largest $\ell_1$ error in the conditional distribution over $\Selected$ is $4.13\times10^{-8}$. Joint optimization includes the conditional tail term in \cref{eq:coarse_observable}, so this test evaluates both components of the population solution. 
	
	\subsection{Gradient Verification}
	
We compare automatic differentiation with the closed-form gradients in float64 arithmetic. The calculations cover full-support distributions and distributions with a vanishing residual mass, and reproduce the equilibrium curves in \cref{fig:mass_equilibrium}.
	
	\section{Response-Length Diagnostics}
	\label{app:legacy_evidence}
	\label{app:legacy_stability}
	\label{sec:stability}
	
	\subsection{Configuration}
	
	The long-run diagnostic uses a mass-preserving variant with an entropy gate, a forward loss averaged over the selected tokens, and a clipped PPO-style reverse objective. Table~\ref{tab:integration_config} gives its training configuration.
	
	\begin{table}[ht]
		\centering
		\normalsize
		\caption{Training configuration for the $\beta=0$ response-length experiment.}
		\label{tab:integration_config}
		\begin{tabular}{ll@{\qquad}ll}
            \toprule
            Setting & Value & Setting & Value \\
            \midrule
            Student & 0.6B & Teacher & 8B \\
            Training examples & 144,490 & Optimizer steps & 2,257 \\
            Learning rate & $3\times10^{-6}$ & Rollout batch & 64 \\
            Samples / prompt & 8 & Response cap & 4,096 \\
            $\alpha$ & 1 & $\beta$ & 0 \\
            \bottomrule
        \end{tabular}
	\end{table}
	
	\subsection{Effect of Residual-Mass Trimming}
	
	Increasing $\beta$ reduces the forward target's residual mass from $1-m$ to $(1-\beta)(1-m)$. When the end-of-sequence token belongs to the residual set, its probability is part of this aggregate target, so residual trimming may also affect when the model stops.
	
	Short-run diagnostics use a 0.6B student, an 8B teacher, and a 4,096-token response budget. With the remaining settings matched, $\beta=0$ keeps response length near 470 for 50 steps, whereas $\beta=0.5$ exhibits response-length degeneration. Because $\beta$ also changes the conditional weight $\rbeta$, this comparison does not isolate the effect of residual-mass trimming on stopping behavior.
	
	\subsection{Long-Run Behavior}
	
	The $\beta=0$ run maintains a mean response length of approximately 478 tokens and a response truncation rate below 0.5\% (Table~\ref{tab:stability}).
	
	\begin{table}[t]
		\centering
		\normalsize
		\caption{\textbf{Response-length behavior at $\beta=0$.} Values are training diagnostics; response truncation denotes the fraction of responses reaching the generation limit.}
		\label{tab:stability}
		\begin{tabular}{lrrr}
			\toprule
			Metric & Full-run mean & Last 100 & Final \\
			\midrule
			Response length & 477.878 & 470.832 & 482.027 \\
			Response truncation (\%) & 0.461 & 0.441 & 1.172 \\
			Reverse loss & 1.032 & 0.948 & 0.971 \\
			Mass-preserving forward loss & 1.046 & 0.937 & 1.048 \\
			Teacher top-$k$ mass & 0.999742 & 0.999713 & 0.999664 \\
			Student top-$k$ mass & 0.993973 & 0.9993896 & 0.993832 \\
			\bottomrule
		\end{tabular}
	\end{table}
	
\section{Experimental Details}
\label{app:large_protocol}

\subsection{Benchmarks and Evaluation}
\label{app:benchmark_protocol}
\label{app:large_reporting}

Table~\ref{tab:benchmark_protocol} lists the evaluation datasets and splits. AMC uses the 83-problem AIMO validation set adapted from AMC12 2022--2023~\citep{li2024numinamath}.

\begin{table}[t]
\centering
\normalsize
\caption{\textbf{Evaluation benchmarks.} HumanEval+ and MBPP+ use the full extended test suites.}
\label{tab:benchmark_protocol}
\begin{tabular*}{\linewidth}{@{\extracolsep{\fill}}lrl@{}}
\toprule
Benchmark & Items & Task / split \\
\midrule
MATH500 & 500 & Mathematics test subset \\
OlympiadBench & 675 & English, text-only math \\
AMC & 83 & AMC12 2022--2023, integer answers \\
AIME24 & 30 & 2024 AIME I and II \\
AIME25 & 30 & 2025 AIME I and II \\
HumanEval+ & 164 & EvalPlus Python tasks \\
MBPP+ & 378 & EvalPlus sanitized tasks \\
ARC-C & 1,172 & Challenge test split \\
MMLU-Pro & 12,032 & Official test split \\
GPQA-Diamond & 198 & Diamond subset \\
\bottomrule
\end{tabular*}
\end{table}

\paragraph{Mathematical reasoning.}
We generate eight responses per problem at temperature 0.7 and top-$p=0.95$, with no top-$k$ truncation and a maximum of 8,192 generated tokens. Each response is scored independently using final-answer extraction and the benchmark's answer checker. For benchmark $b$ with $N_b$ problems, let $c_i$ be the number of correct responses among eight samples for problem $i$. We report their mean correctness as avg@8
\begin{equation}
    \operatorname{avg@8}(b)=\frac{100}{N_b}\sum_{i=1}^{N_b}\frac{c_i}{8}.
\end{equation}

\paragraph{Code generation.}
HumanEval+ and MBPP+ use zero-shot greedy decoding with one completion per task, top-$p=1$, and an 8,192-token response limit. We score completions with the extended EvalPlus test suites.

\paragraph{Out-of-domain evaluation.}
ARC-C, MMLU-Pro, and GPQA-Diamond are evaluated on the same epoch-1 mathematics-distilled checkpoints used for the mathematics benchmarks. We evaluate MMLU-Pro using the official five-shot chain-of-thought protocol on the full test set. Demonstrations are drawn from the validation split of the same subject. The official evaluation configuration uses greedy decoding, a 2,048-token output limit, and the stop string \texttt{Question:}; answer extraction and accuracy follow the official evaluator~\citep{wang2024mmlupro}. ARC-C and GPQA-Diamond use zero-shot generated option labels, with the same prompts and option order across methods. Invalid or missing labels count as incorrect on these two benchmarks. These evaluations measure generated-answer accuracy rather than option log-likelihood ranking.

\subsection{Training-Dynamics Diagnostic}
\label{app:training_diagnostics}

\begin{figure}[t]
    \centering
    \includegraphics[width=0.66\linewidth]{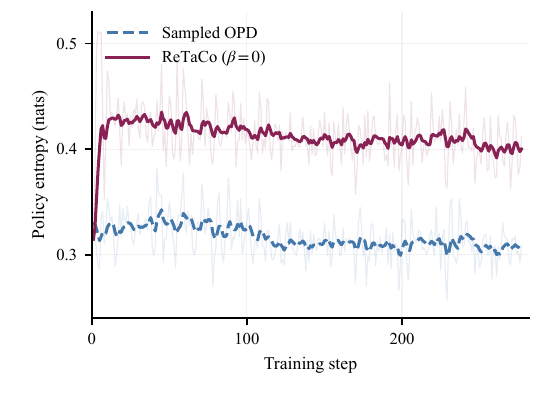}
    \caption{\textbf{Policy entropy during training.} Qwen3-1.7B student, Qwen3-30B-A3B-Instruct-2507 teacher, and $k=16$. The curves show \texttt{actor/entropy} in nats for Sampled OPD (blue, dashed) and \method\ with $\beta=0$ (wine, solid), over the common interval of steps 1--277. Solid and dashed curves use exponential smoothing with span 11; faint traces show raw values. The last 20 steps average 0.305 and 0.399 nats, respectively.}
    \label{fig:policy_entropy}
\end{figure}

\paragraph{Policy entropy.}
In this run, \method\ with $\beta=0$ maintains higher policy entropy than Sampled OPD (Figure~\ref{fig:policy_entropy}). Both runs use the Qwen3-1.7B student, the Qwen3-30B-A3B-Instruct-2507 teacher, and $k=16$, and we compare them over their common interval of steps 1--277. Over the last 20 steps, the recorded \texttt{actor/entropy} averages 0.399 nats for \method\ and 0.305 nats for Sampled OPD, and the two trajectories remain separated after the initial transient.

\begin{figure}[t]
    \centering
    \begin{subfigure}[t]{0.49\linewidth}
        \centering
        \includegraphics[width=\linewidth]{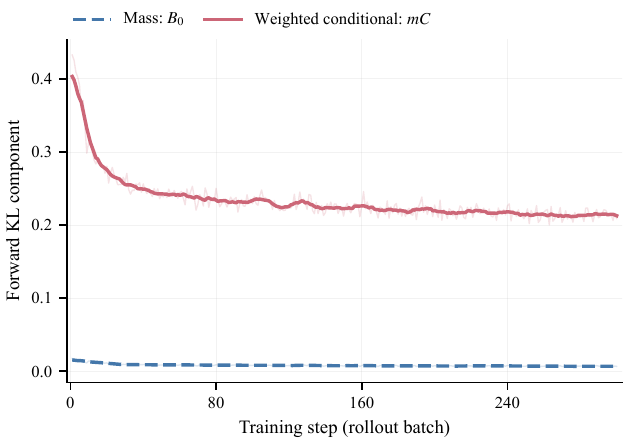}
        \caption{Forward-loss decomposition.}
    \end{subfigure}\hfill
    \begin{subfigure}[t]{0.49\linewidth}
        \centering
        \includegraphics[width=\linewidth]{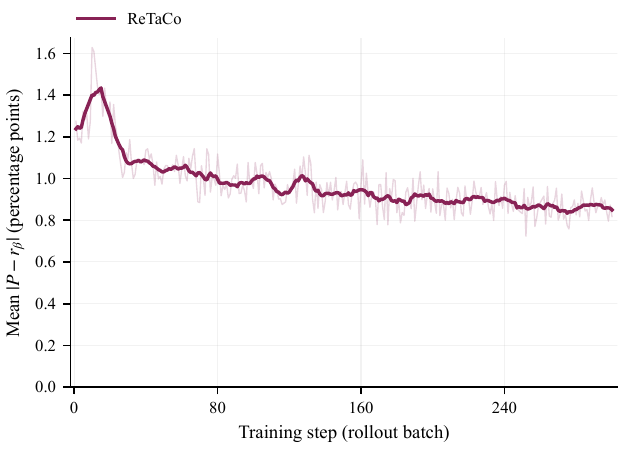}
        \caption{Absolute mass error.}
    \end{subfigure}
    \caption{\textbf{Mass and conditional matching during training.} A single Qwen3-1.7B mathematics training run with $k=16$, $\alpha=1$, and $\beta=0$. (a) Mean $B_0=\BerKL(m\|P)$ and $mC=m\KL(\qsel\|\psel)$ from \cref{eq:coarse_forward}. (b) Mean $|P-m|$ in percentage points. Faint lines show raw values; dark lines show centered 11-step moving averages, not confidence intervals. Steps denote rollout batches. This diagnostic uses a 7,168-token response cap and a 300-step budget, distinct from the main benchmark protocol.}
    \label{fig:training_decomposition}
\end{figure}

\paragraph{Additional loss decomposition.}
Both the conditional component of the forward loss and the absolute mass error decline during a Qwen3-1.7B mathematics training run with the Qwen3-30B-A3B-Instruct-2507 teacher (Figure~\ref{fig:training_decomposition}). This seed-11 run comprises 300 consecutive rollout batches with a 7,168-token response limit, $k=16$, $\alpha=1$, and $\beta=0$. It is a separate diagnostic from the one-epoch benchmark and ablation comparisons. The curves summarize response-token statistics using micro-batch aggregation, which differs from the reduction used in the training loss.

The weighted conditional component is computed as the mean forward KL minus the mean Bernoulli KL, preserving the identity in \cref{eq:coarse_forward}; it is not the product of separately averaged mass and conditional KL. Absolute mass error is the mean of per-token absolute deviations, not the absolute value of a signed mean. Initial and final values are averaged over the first and last 20 steps, respectively. For display only, an 11-step centered moving average uses the available neighborhood at each endpoint. No run averaging or uncertainty intervals are shown.

Numerical tail projection is active at an average of 31.65\% of response-token positions over the last 20 steps. These curves therefore describe the numerically stabilized implementation on evolving student-generated prefixes, rather than directly measuring the population optimum at a fixed prefix without clipping. The declines in conditional loss and mass error are descriptive only; the component ablations in \cref{sec:ablations} assess how each term affects downstream accuracy.

\subsection{Models and Optimization}

Teacher and student models in each pair (Table~\ref{tab:large_main}) share compatible token vocabularies. The teacher is frozen and scores student-generated prefixes.

\paragraph{Training data.}
Mathematics training uses DAPO-Math-17k~\citep{yu2025dapo}. Code-domain distillation uses the code subset of Eurus-2-RL-Data~\citep{cui2025process}, with separate mathematics and code training runs initialized from the original student. Only problem prompts enter the distillation procedure; reference solutions and correctness rewards are not used. Each code experiment uses 8$\times$ NVIDIA H200 GPUs.

\begin{table}[t]
\centering
\begingroup\normalsize
\setlength{\tabcolsep}{3pt}
\caption{\textbf{Mathematics training settings shared by Qwen3-1.7B, Qwen3.5-2B, and Gemma4-E2B.} $H(q)$ is full-vocabulary teacher entropy.}
\label{tab:training_config}
\begin{tabular}{ll}
\toprule
Setting & Value \\
\midrule
Optimizer / learning rate & AdamW / $3\times10^{-6}$ \\
AdamW betas / epsilon & $(0.9,0.999)$ / $10^{-8}$ \\
Weight decay & 0 \\
Schedule / warmup & Cosine / 3\% \\
Precision / gradient clipping & BF16 / 1.0 \\
Prompt batch / mini-batch & 128 / 32 \\
Micro-batch per GPU & 1 \\
Rollouts per prompt & 1 \\
Optimization passes per batch & 1 \\
Prompt / response limits & 2,048 / 8,192 tokens \\
Training temperature / top-$p$ & 1.0 / 1.0 \\
Training top-$k$ truncation & Disabled \\
Repetition penalty & 1.0 \\
Chat template / thinking & Native student / disabled \\
\method\ $k$, $\alpha$, $\beta$ & 16, 1, 0 \\
EOPD $k$, $\alpha$, $\beta$ & 16, 1, 1 \\
\method\ forward gate & None \\
EOPD forward gate & $H(q)>0.8$ \\
Loss reduction & Response-token mean \\
Checkpoint epoch & 1 \\
\bottomrule
\end{tabular}
\endgroup
\end{table}

\paragraph{Optimization and objectives.}
\Cref{tab:training_config} lists the training settings. The Qwen3 setup uses full student fine-tuning with gradient checkpointing. Sampled OPD and \method\ use the direct $k3+$ estimator in verl~\citep{sheng2024hybridflow}, with losses averaged over unmasked response tokens. No task reward, reference-policy penalty, or entropy bonus is added.

\paragraph{Model-specific settings.}
The Qwen3.5 setup uses 16 H200 GPUs, with eight each for student and teacher in Sampled OPD and \method; EOPD colocates its models across all 16 GPUs.

\paragraph{EOPD baseline.}
EOPD uses OpenRLHF with the forward target and entropy gate in \cref{tab:training_config}, normalized over all unmasked response tokens. Its reverse objective is a PPO-style surrogate with importance-ratio clipping to $[0.8,1.2]$; its advantages are the detached differences between teacher and behavior log-probabilities. Behavior probabilities are fixed within each rollout batch.

\section{Additional Ablations}
\label{app:extended_ablations}

\subsection{Component Contributions}

All ablations use the Qwen3 teacher--student pair and epoch-1 checkpoints, with evaluation as in \cref{app:benchmark_protocol}.

\begin{table}[t]
\centering
\normalsize
\caption{\textbf{Response truncation in the component ablations.} Each variant trains for one epoch. Truncation denotes the fraction of responses reaching the 8,192-token generation limit.}
\label{tab:ablation_full}
\begin{tabular}{@{}llr@{}}
\toprule
Variant & Objective & Truncated (\%) \\
\midrule
Mass-only forward term & $R+B_0$ & 47.51 \\
Conditional-only forward term & $R+mC$ & 48.33 \\
No reverse term & $B_0+mC$ & 49.98 \\
\bottomrule
\end{tabular}
\end{table}

All three component variants reach the response limit on nearly half of their completions (Table~\ref{tab:ablation_full}).

\subsection{Hyperparameter Sensitivity}

With $k=16$ and $\beta=0$, the default $\alpha=1$ outperforms $\alpha=0.5$ on all five tasks (Table~\ref{tab:ablation_sweeps}). Increasing $\alpha$ to 2 reduces MATH500 from 83.78\% to 70.45\% and AIME24 from 28.75\% to 25.00\% but raises AIME25 from 21.67\% to 22.50\%, so a larger forward weight does not improve all benchmarks uniformly.

Increasing $k$ from 8 to 32 raises AIME24 from 23.33\% to 27.50\% but lowers AIME25 from 22.50\% to 18.33\%, so the effect of $k$ also varies across tasks.

\begin{table}[t]
\centering
\caption{\textbf{Qwen3-1.7B hyperparameter sensitivity (avg@8, \%).} Every setting trains for one epoch. Defaults are $k=16$, $\alpha=1$, and $\beta=0$; each group varies the indicated parameter.}
\label{tab:ablation_sweeps}
\begingroup\normalsize
\setlength{\tabcolsep}{3pt}
\begin{tabular*}{\linewidth}{@{\extracolsep{\fill}}lccccc@{}}
\toprule
Setting & MATH500 & OlympiadBench & AMC & AIME24 & AIME25 \\
\midrule
\multicolumn{6}{l}{\textit{Forward weight}} \\
$\alpha=0.5$ & 78.45 & 43.78 & 46.84 & 21.25 & 19.17 \\
$\alpha=1$ (default) & 83.78 & 49.65 & 52.56 & 28.75 & 21.67 \\
$\alpha=2$ & 70.45 & 44.15 & 47.59 & 25.00 & 22.50 \\
\midrule
\multicolumn{6}{l}{\textit{Selected support}} \\
$k=8$ & 79.25 & 45.76 & 49.85 & 23.33 & 22.50 \\
$k=32$ & 79.35 & 46.17 & 49.10 & 27.50 & 18.33 \\
\bottomrule
\end{tabular*}
\endgroup
\end{table}

	
	\section{Related Work}
	\label{sec:related_work}
	
	\paragraph{Knowledge distillation and divergence choice.}
	Knowledge distillation transfers teacher behavior through soft targets or generated sequences~\citep{hinton2015distilling,kim2016sequence}. Forward KL emphasizes tokens the teacher supports, whereas reverse KL emphasizes tokens the student already favors. MiniLLM adopts reverse KL~\citep{gu2024minillm}, generalized knowledge distillation (GKD) studies divergence choice and trajectory mixtures~\citep{agarwal2024policy}, and other methods balance the two directions globally or token by token~\citep{amara2022bd,wu2025rethinking,jung2025todi}. Our analysis examines how compressing and renormalizing the teacher changes target mass even when the divergence direction is fixed.
	
	\paragraph{On-policy language-model distillation.}
	On-policy distillation queries teachers at student-generated prefixes to reduce state mismatch~\citep{agarwal2024policy,lu2025onpolicydistillation}. EOPD adds forward KL at positions with high teacher entropy to recover teacher-supported tokens that sampled reverse KL may underweight~\citep{jin2026eopd}. \method\ shares EOPD's token-recovery motivation but separates the conditional shape of the selected tokens from their aggregate mass. EOPD uses entropy gating to select supervised positions; the default \method\ objective applies its explicit residual target without an entropy gate.
	
	\paragraph{Compressed teacher distributions.}
	Sparse teacher targets reduce the communication and storage costs of dense teacher distributions~\citep{shum2024first,peng2025pre}, and tail-aware distillation separates top-$k$ predictions from their lower-probability tail~\citep{dasgupta2026tail}. Our work differs in three ways: we diagnose EOPD's renormalized forward term exactly, we pair residual aggregation with a single-sample estimator whose expectation equals the full-vocabulary reverse KL, and we parameterize the residual target. The analysis characterizes how this target controls the joint optimum across mass preservation and deliberate trimming.
	
	\paragraph{Uncertainty and reasoning diversity.}
	Entropy-aware optimization promotes exploration at uncertain reasoning positions~\citep{cheng2025reasoning,wang2025beyond}. Entropy bonuses control distributional spread, whereas forward distillation supplies teacher-specific alternatives. The recovery gradient for the selected tokens in \method\ follows from teacher-specific forward supervision and remains non-vanishing at $\beta=0$, without requiring a global entropy intervention.

\end{document}